\documentclass[letterpaper, 10 pt, conference]{ieeeconf}  
\usepackage[T1]{fontenc}
\usepackage{booktabs}
\usepackage{cite}

\IEEEoverridecommandlockouts                              

\usepackage{amsmath} 
\usepackage[dvipsnames]{xcolor}
\usepackage{bbding}
\usepackage{pifont}
\usepackage{wasysym}
\usepackage{amssymb}
\usepackage[pdftex]{graphicx}
\usepackage{multicol}
\usepackage{colortbl}
\usepackage{hyperref}
\usepackage{url}
\usepackage{multirow}
\usepackage{tabularx}
\usepackage{mathtools}
\usepackage{xcolor}
\usepackage[nolist]{acronym}
\usepackage{amssymb}
\usepackage{pifont}
\RequirePackage{tikz} 

\begin{acronym}
    \acro{ROV}{remotely operated vehicle}
    \acro{AUV}{autonomous underwater vehicle}
    \acro{UWRS}{underwater robotics simulator}
    \acro{NMPC}{nonlinear model predictive control}
    \acro{MPC}{model predictive control}
    \acro{SLAM}{simultaneous localisation and mapping}
    \acro{vSLAM}{visual SLAM}
    \acro{RL}{reinforcement learning}
    \acro{DRL}{deep reinforcement learning}
    \acro{UE}{unreal engine}
    \acro{NED}{North-East-Down}
    \acro{UE5}{unreal engine 5}
    \acro{UE4}{unreal engine 4}
    \acro{ROS}{robot operating system}
    \acro{AI}{artificial intelligence}
    \acro{APE}{absolute positioning error}
    \acro{RPE}{relative positioning error}
    \acro{PWM}{pulse width modulation}
    \acro{DFKI}{Deutsches Forschungszentrum für Künstliche Intelligenz}
\end{acronym}

\title{\LARGE \bf
UNav-Sim: A Visually Realistic Underwater Robotics Simulator and Synthetic Data-generation Framework
}

\author{ 
Abdelhakim Amer, Olaya Álvarez-Tuñón, Halil \.{I}brahim U\u{g}urlu, Jonas le Fevre Sejersen, \\ Yury Brodskiy and
Erdal Kayacan
\thanks{A. Amer, O. Tunon, H. U\u{g}urlu, J. Sejersen  are with the Department of Electrical Engineering and Computer Engineering, Aarhus University, 8200 Aarhus, Denmark {\tt\small \{abdelhakim, olaya, halil, jonas\} at ece.au.dk}.
Y. Brodskiy is with EIVA a/s, 8660 Skanderborg, Denmark. {\tt\small \{ybr\} at eiva.com}.
E. Kayacan is with the Automatic Control Group, Department of Electrical Engineering and Information Technology, Paderborn University, Paderborn, Germany. {\tt\small \{erdal.kayacan\} at uni-paderborn.de}}
}
\begin{document}

\maketitle
\thispagestyle{empty}
\pagestyle{empty}

\begin{abstract}

Underwater robotic surveys can be costly due to the complex working environment and the need for various sensor modalities. While underwater simulators are essential, many existing simulators lack sufficient rendering quality, restricting their ability to transfer algorithms from simulation to real-world applications. To address this limitation, we introduce UNav-Sim, which, to the best of our knowledge, is the first simulator to incorporate the efficient, high-detail rendering of Unreal Engine 5 (UE5). UNav-Sim is open-source\footnote{\url{https://github.com/open-airlab/UNav-Sim}} and includes an autonomous vision-based navigation stack. By supporting standard robotics tools like ROS, UNav-Sim enables researchers to develop and test algorithms for underwater environments efficiently.
\end{abstract}

\section{Introduction}\label{sec:intro}
Marine robotics is an expanding field with numerous applications, including exploring underwater ecosystems and inspecting underwater infrastructure \cite{applications}. Recent developments in robotics and autonomy have demonstrated the superior capabilities of \ac{AI} and vision-based algorithms in solving complex tasks, such as drone racing \cite{DRL_erdal,gatenet} and inspection \cite{mohit}. These achievements have shown promise in developing \ac{AI} and autonomy for marine applications as well \cite{DFKI}. To mitigate the high costs involved in developing and testing such algorithms, photorealistic simulation environments are needed that can accurately model the complexity of underwater scenarios \cite{mimir23}.

\begin{figure}[t]
    \centering
    \includegraphics[width=8.5cm]{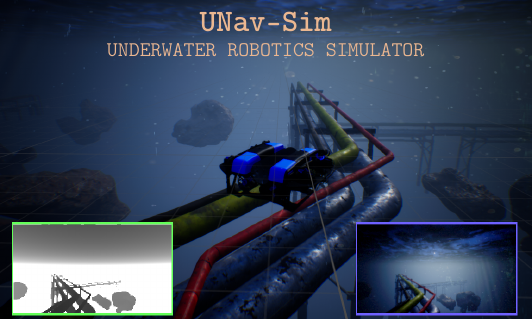}
    \caption{UNav-Sim is an underwater robotics simulator utilizing Unreal Engine 5 (UE5) highly realistic environments. The simulator includes many features useful for roboticists, such as ROS 2, and a wide range of sensors and cameras. The bottom right displays the feed from a front-facing RGB camera, while the bottom left shows a corresponding depth image.}
    \label{fig:uwrs}
\end{figure}

\begin{figure*}[!t]
    \centering
    \includegraphics[width=0.97\textwidth]{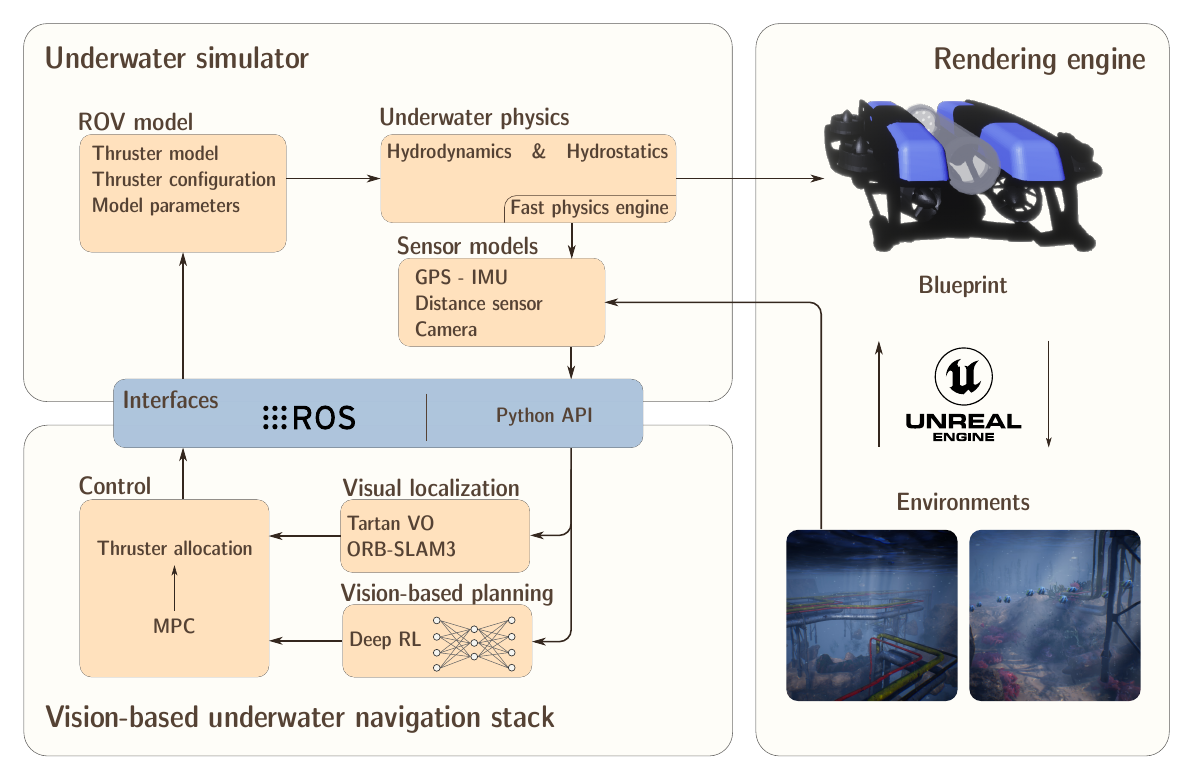}
    \caption{UNav-Sim system architecture is designed to be modular, allowing flexibility in adapting the simulator to various underwater autonomy tasks. The system utilizes Unreal Engine 5 (UE5) to provide a high-fidelity rendering environment for increased photo-realism.  A model predictive controller (MPC), combined with a deep reinforcement learning (DRL) planner and Visual \ac{SLAM} are utilized for vision-based underwater navigation.   }
    \label{fig:abstract} 
\end{figure*}

In this context, this paper presents UNav-Sim, the first open-source underwater simulator based on \ac{UE5} to create photorealistic environments (see Fig. \ref{fig:uwrs}). Compared to existing underwater robotics simulators, \cite{holoocean,uuv,dave,marus}, UNav-Sim provides superior rendering quality, essential for the development of \ac{AI} and vision-based navigation algorithms for underwater vehicles. It supports robotics tools such as ROS 2 and autopilot firmwares making it suitable for robotics research and development. The simulator uses the following open-source AirSim \cite{airsim} extensions: \cite{byu_vtol} to add custom vehicle models to AirSim, and \cite{coles} for integration of AirSim to \ac{UE5}. UNav-Sim can be used to simulate a wide range of underwater scenarios and models. The paper also demonstrates its effectiveness for the development of vision-based localization and navigation methods for underwater robots. 

The rest of the paper is structured as follows: Section \ref{sec:soa} presents an overview of the state-of-the-art simulators and their respective capabilities. Section \ref{sec:description} describes the software architecture, physics, and models that comprise UNav-Sim. In Section \ref{sec:stack}, we describe a vision-based underwater navigation stack that was developed as a component of UNav-Sim. Then, we present a test case in Section \ref{sec:tests}, where we showcase the abilities and features of our simulator in a vision-based pipe inspection scenario. Lastly, conclusions are drawn from this work in Section \ref{sec:conclusion}.

\section{State-of-the-art} \label{sec:soa}

Robotics simulation tools have significantly advanced in recent years, with a focus on providing  high-fidelity and photorealistic visual rendering. IsaacSim \cite{isaac}, developed by Nvidia, is a recent example that includes both high-fidelity contact simulation and high-quality image rendering provided by Omniverse, making it suitable for simulating robotic grippers and walking robots. Another example is Microsoft's AirSim \cite{airsim}, yet another popular robotics simulator, specifically designed for aerial vehicles. AirSim utilizes its Fastphysics engine for physics simulation and \ac{UE4} for visualization.

While progress in robotics simulation tools has been rapid, underwater robotics simulation tools have lagged behind. UWSim \cite{uwsim} and UUV Simulator \cite{uuv} are the two most commonly used underwater simulators \cite{survey}; however, they are now discontinued. 
A more recent simulator, DAVE \cite{dave}, was developed as a more modern version of the UUV simulator that supports more \ac{ROV} models and underwater sensors. However, the aforementioned simulators are based on Gazebo, which has the disadvantage of unrealistic rendering. This limits their usefulness for training and testing \ac{AI} algorithms that often rely on image inputs. To address this issue, HoloOcean \cite{holoocean} was developed using \ac{UE4} for rendering and written in Python, but it lacks support for \ac{ROS} \cite{ros}. Another example is MARUS \cite{marus}, which has not yet released its open-source implementation. The simulator uses Unity3D for visualization and integrates with \ac{ROS}. However, it lacks support for essential robotics and \ac{AI} tools, such as commercial autopilots \cite{px4}, or OpenAI's Gym environments \cite{gym}, which are important tools for developing \ac{AI} and control algorithms for autonomous vehicles.
\begin{table*}[t]
\caption{Marine robotics simulators comparison showing UNav-Sim's superior rendering quality and versatility.}
\centering

\begin{tabular}{ cccccc}

\toprule


Simulator & Year & Rendering  quality &ROS Support& Autopilot & OS 
\\
\midrule

UWSim \cite{uwsim} & 2012 &Low& ROS 1 & None & Linux \\

UUV \cite{uuv} &2016 & Low& ROS 1 & Ardupilot & Linux 
 \\

URSim \cite{ursim} & 2019 & Moderate & ROS 1 & N/A& Linux \\

HoloOcean \cite{holoocean} & 2022 & High  & N/A& N/A  & Linux/Windows  \\

DAVE \cite{dave}& 2022 &Low &  ROS 1& PX4/Ardupilot  &Linux  \\

MARUS \cite{marus} & 2022 & Moderate & ROS 1,2 & N/A & Linux/Windows
  \\
\midrule

\textbf{UNav-Sim (Ours)} &2023 & Highest & ROS 1,2 & PX4/Ardupilot &  Linux/Windows \\
\bottomrule

\label{Tab:comparision}

\end{tabular}

\label{Tab:comparision2}
\end{table*}

A comparison of the capabilities of various open-source underwater robotics simulators, including the proposed simulator, is presented in Table \ref{Tab:comparision}. Amongst all simulators evaluated, the present work, UNav-Sim, stands out for its superior rendering quality, achieved through the utilization of the \ac{UE}5 graphics engine. Additionally, UNav-Sim supports a range of tools commonly used in developing robotics solutions, such as \ac{ROS}, gym environments, and autopilot systems. Furthermore, UNav-Sim is compatible with both Windows and Linux operating systems.

\section{UNav-Sim software architecture} \label{sec:description}

UNav-Sim is composed of three main components, as illustrated in Fig.~\ref{fig:abstract}: an underwater physics simulator, a state-of-the-art rendering engine, i.e. \ac{UE5}, and an autonomy stack. The underwater physics simulator, which contains the lumped parameter \ac{ROV} model and underwater dynamics equations, is modular and allows underwater vehicle motion simulation. 
It leverages the capabilities of AirSim, including the Fastphysics solver and a range of sensor models, such as GPS, IMU, cameras, and distance sensor. 
An API allows communication between the navigation stack and the physics simulator, with the former receiving essential sensor data and sending control commands. A \ac{ROS} wrapper is also available, which enables \ac{ROS}-based development and communication between different modules.


\subsection{Underwater environment rendering}

Underwater image formation can be modelled as a superposition of absorption, forward scattering, and backscattering effects at each pixel $\textbf{x} = (u, v)$. The image intensity $I_c(x)$ in each color channel $c$ can be expressed as \cite{olaya}:

\begin{equation}\label{eq:lighting_model}
I_c(x) = D_c(x) + F_c(x) + B_c(x)
\end{equation}

In this equation, $D_c$ represents the attenuated signal from the object due to absorption. The forward scattering component $F_c$ captures the light from the object that reaches the camera with small-angle scattering. Lastly, the backscattering component $B_c$ accounts for the degradation in color and contrast caused by the water scattering effect, where the light does not originate directly from the object. These effects can be modelled using different techniques and can vary based on the implementation within the rendering engine.

UNav-Sim utilizes \ac{UE5} as the rendering engine, which offers significant improvements over its predecessor, \ac{UE4}. \ac{UE5} significantly boosts polygon handling to 10 billion, introduces real-time ray-based lighting with Lumen, and incorporates Temporal Super Resolution for high-quality textures with minimal performance impact, enhancing visual fidelity and efficiency. 


\ac{UE5} underwater rendering module models scattering effects in underwater images \eqref{eq:lighting_model}, with Schlick Phase Functions \cite{schlick}, taking into account the Opaque or Masked water surface. 
The transparency of the water is implicitly handled within the volume shading model, 
and refraction is managed by reading the depth and color beneath the water surface to distort the samples.  
One of the main challenges in generating underwater renderings is the variety of imaging conditions that drastically change the environment's appearance \cite{akkaynak2017space}. \ac{UE5} allows users to define the scattering coefficients, absorption coefficients, phase function, and color scale behind the water, providing control over the water's appearance and thus allowing users to simulate their preferred environment's conditions. 

Consequently, using \ac{UE5} within UNav-Sim, underwater environments that appear realistic can be created, where \ac{UE} allows designers to place and manipulate assets in a 3D space. 
These assets can include terrain, static meshes, and lighting. They can be customized to create underwater virtual worlds, as shown in Fig. \ref{fig:abstract}. 

UE uses blueprints to define the physical representation and behaviour of an \ac{ROV}. In UNav-Sim, the blueprint is linked to an external underwater physics engine to obtain kinematic information. The blueprint also defines cameras that gather visual information from the underwater environment, such as RGB and depth images.

\subsection{Underwater physics}

The core of the physics underlying underwater vehicles consists of the equations of motion that describe the different forces and moments acting on the vehicle's body. These forces and moments can be classified into three categories: hydrostatics, hydrodynamics, and externally applied forces.



\begin{figure}[!t]
    \centering
    \includegraphics[width=\columnwidth]{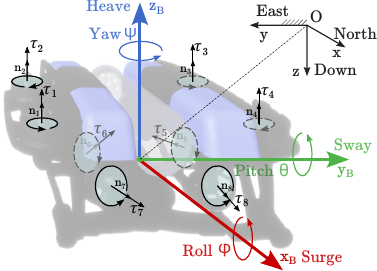}
    \caption{ROV model defined in UNav-Sim: A force $\tau_i$ is applied at each thruster location $i$, where the thruster orientation is defined by a vector $\textbf{n}_i$.}
    \label{fig:rov_fbd}
\end{figure}

The equation of motion in the body-fixed frame, originally presented in Fossen \cite{fossen}, can be expressed in SNAME notation \cite{SNAME} as, 
\begin{equation}\label{eq:eom}
  \begin{multlined}
    \mathbf{M}_{RB}\dot{\mathbf{\nu}}= \tau - \underbrace{ \mathbf{C}_{RB}(\mathbf{\nu})\mathbf{\nu} }_\text{Coriolis \hspace{1 pt} term} - \underbrace{ \mathbf{M}_A \dot{\mathbf{\nu}} - 
    \mathbf{C}_A(\mathbf{\nu}) \mathbf{\nu} }_\text{Added\hspace{2pt}mass}     
    \\
    - \underbrace{ \mathbf{D}(\mathbf{\nu})\mathbf{{\mathbf{\nu}} }}_\text{Drag } - \mathbf{g}(\mathbf{\eta}).
    \end{multlined}
\end{equation}
The vehicle's pose, $\mathbf{\eta}=[x, y, z, \phi, \theta, \psi]^T$, is described by a six-dimensional column vector where $x$, $y$, and $z$ denote the vehicle's position in the \ac{NED} frame, while $\phi$, $\theta$, and $\psi$ represent its roll, pitch, and yaw angles, respectively (see Fig. \ref{fig:rov_fbd}).
The linear and angular velocity vector in the body-fixed frame is denoted as $\mathbf{\nu} = [u, v, w, p, q, r]^T$. The inertia matrix of the vehicle's body is represented by $\mathbf{M}_{RB}$.
Hydrostatic forces, $g(\eta)$, arise from gravity and buoyancy, along with associated moments and torques.
Hydrodynamic forces arise from the interaction between the vehicle and the surrounding water and can significantly impact the vehicle's behavior. These forces include the Coriolis and centripetal forces caused by the rigid body's mass, $\mathbf{C}_{RB}(\mathbf{\nu})\mathbf{\nu}$, the Coriolis forces, $\mathbf{C}_A(\mathbf{\nu})$, and moment of inertia, $\mathbf{M}_{A}$, arising from the added mass, and linear and quadratic damping effects, $\mathbf{D}(\mathbf{\nu})\mathbf{{\nu} }$.
External forces, $\tau$, include the forces exerted on the \ac{ROV} by its thrusters, as well as the disturbances, caused by the surrounding water flow. Multiple disturbance models, such as constant value, sinusoidal, and a combination of sine waves are implemented.

To efficiently solve the equations of motion presented above, AirSim's high-frequency physics engine is utilized with a computational frequency of $1000 \textrm{Hz}$. The engine uses the velocity verlet algorithm for numerical integration due to its computational benefits.


\subsection{ROV model }


The \ac{ROV} is represented as a rigid body that is manipulated by an arbitrary number of actuators, $N$. These actuators are located at user-defined vertices of the vehicle, with corresponding normals and positions denoted by $n_i$ and $r_i$, respectively, where $i \in \{1,\dots, N\}$ is the actuator number. 
At each vertex, the vehicle receives a unitless control input, $u_i\in (-1,1)$. To account for actuator dynamics, a discrete low-pass filter with a time constant of $t_c$ is applied to the control input. The filtered input, denoted as $u_{fi}$, is then used to calculate the thrust force using the relationship given as \cite{airsim},
\begin{equation}\label{eq:thruster_model}
  \tau_i = C_T \rho \omega^2_{max} D^4 u_{fi}, 
\end{equation}
where $\tau_i$ is the thrust force on the $i$-th thruster, $C_T$ is the thrust coefficient, $\rho$ is the density of water, $\omega_{max}$ is the maximum thruster rotation speed, and $D$ is the propeller diameter. 
In order to accurately compute the specific rigid motion of the \ac{ROV}, as described by equations \eqref{eq:eom} and \eqref{eq:thruster_model}, several vehicle-specific parameters such as its inertia, hydrodynamic coefficients, and the maximum thruster rotation speed, must be configured by the user. 

The model chosen to be implemented, the Blue Robotics BlueROV2 Heavy, is an over-actuated \ac{ROV} with four vertical thrusters and four horizontal thrusters as shown in Fig. \ref{fig:rov_fbd}. The horizontal thrusters are oriented at 45 degrees and are responsible for the control of three degrees of freedom, namely, surge, sway, and yaw, while the vertical thrusters control heave, pitch, and roll. The model parameters are obtained from \cite{bluerov2_h}.


\section{Vision-based underwater navigation stack}\label{sec:stack}
The vision-based underwater navigation stack of UNav-Sim includes planning, control, and \ac{SLAM}. Many state-of-the-art \ac{AI} algorithms in robotics, for example, as implemented in OpenDR toolkit \cite{opendr} or PyPose library \cite{pypose}, primarily rely on vision for localization, planning, and control~\cite{huy}. Therefore, in this study, we have chosen to utilize visual SLAM (VSLAM) and end-to-end \ac{DRL} algorithms to demonstrate the capabilities of the proposed simulator. To facilitate ease of integration with various autonomy algorithms and deployment on actual hardware, all algorithms have been developed using \ac{ROS} framework.

\subsection{Vision-based planning}
\label{sec:stack:planning}


The recent developments in machine learning methods enable intelligent agents to learn navigation tasks end-to-end. Generally, a deep neural network policy takes sensory input, such as high-dimensional visual data, and generates feasible actions without explicitly mapping the environment. These learning-based methods require large amounts of data for training, which is impractical for real-world robotics systems. Hence, simulation environments are very substantial for enriching the required data in many cases \cite{loquercio2021learning}.
Furthermore, \ac{DRL} methods require demonstrations for exploration of the environment to learn a policy, which increases safety concerns for real-world learning \cite{halil}. 
Consequently, simulation environments with high-fidelity visual sensors and accurate physical dynamics are crucial for DRL research. OpenAI's gym~\cite{gym} is a general standard for experimenting with learning-based sequential decision-making tasks.
Therefore, we have provided a gym environment along with our simulator to augment its capabilities for benchmarking learning-based navigation algorithms.

\subsection{Control}
\label{sec:stack:control}

A \ac{MPC} strategy was utilized to control the \ac{ROV} model, which consists of a two-step process involving an \ac{MPC} followed by a control allocation algorithm. The motivation for the use of MPC is based on its ability to handle both input and state constraints explicitly, and intuitive tuning parameters \cite{mpc_survey}.   
The \ac{MPC} is designed to solve an online optimization problem, aiming to determine the optimal body wrench forces and moments, given a particular robot pose and a desired reference pose. 

The control allocation algorithm is then employed to obtain the individual control signal for each thruster by using a pseudo-inverse of an allocation matrix, which is vehicle-specific and depends on the thruster configuration of the \ac{ROV}. ACADO toolkit \cite{acado} is utilized to implement the \ac{MPC} as a  \ac{ROS} package, which is integrated into the  \ac{ROS} navigation stack.

\subsection{Visual localization}
Visual localization algorithms rely on cameras to retrieve the robot's state, which are widely used in the underwater robotics community \cite{olaya_survey}. 
Long-standing ROS packages for SLAM include \texttt{robot\_localization}, which presents a classical filtering approach for sensor fusion \cite{vSLAM:moore2016robot_localizationROS}, \texttt{hector\_slam}, which implements occupancy grid maps for laser and IMU data \cite{vSLAM:hectorSLAM}, and \texttt{gmapping}, which leverages a Rao-Blackwellized particle filter for laser-based SLAM  \cite{vSLAM:gmapping}.
The availability of these state-of-the-art and ready-to-use algorithms has allowed outstanding progress in the robotics community \cite{vSLAM:lidarreview,vSLAM:hectorreview1,vSLAM:hectorreview2,vSLAM:gmappingreview1,vSLAM:gmappingreview2}, as they ease the implementation of future advances for SLAM and the benchmarking of their performance. However, the availability of off-the-shelf packages for visual SLAM remains an open problem.
Therefore, we propose \texttt{robot\_visual\_localization}, a ROS metapackage for deploying and benchmarking visual localization algorithms. We chose to implement the state-of-art methods ORB-SLAM3 \cite{vSLAM:orb3} and TartanVO \cite{vSLAM:tartanvo}. Each algorithm is implemented as a standalone ROS package within the \texttt{robot\_visual\_localization} metapackage, which takes as input the image stream, and outputs the camera trajectory and the map points for ORB-SLAM3, and the camera trajectory for TartanVO. 

ORB-SLAM3 encompasses a geometry-based approach for SLAM. The ORB-SLAM3's front end tracks ORB features across consecutive frames. The features are selected to be uniformly distributed across the image, and the matches search is performed according to a constant velocity model. The ORB-SLAM3's back end builds a map with the sparse points tracked from the front end. Under tracking loss, the map is stored in memory as inactive, creating a new active map. The loop closure thread finds revisited areas under the active and inactive maps, merging them and propagating the accumulated drift. 

Geometry-based algorithms such as ORB-SLAM3 still present the de-facto state-of-art for SLAM, due to their high precision and efficiency. However, they are highly dependent on feature detection and matching, and therefore sensitive to visual degradations such as repetitive patterns, textureless environments, and non-Lambertian surfaces. On the other hand, learning-based algorithms can be more robust against those challenging imaging conditions, but are highly dependent on the training data, and usually lack generalization ability.

The proposed ROS metapackage serves then as an accessible tool for the easy comparison of two of the main taxonomies in the SLAM's state-of-art, which in the present work serves as a comparison of their performance under challenging underwater imaging conditions.

\section{Example use-cases}\label{sec:tests}
As a case study for UNav-Sim, we present an autonomous pipe inspection scenario. 
Pipe inspection, being the most common use case for \ac{ROV}s, presents a relevant and challenging use case, where vision-based navigation is essential to achieve the required task. Furthermore, we assess the efficacy of our underwater autonomy stack and report its performance in executing the designated autonomous pipe inspection task. A video showing the pipe inspection demonstration can be found here\footnote{\url{https://youtu.be/unZS33lCqpU}}.

\subsection{Vision-based pipe following with DRL}\label{sec:example:planning}

In this experiment, the performance of UNav-Sim is evaluated in a pipe-following task. An agent utilizing \ac{DRL} is trained with the gym interface provided by the simulator to generate position commands based on RGB image observations. An \ac{MPC} controller subsequently executes the position commands. The convolutional neural network policy inputs $180\times320$ pixel RGB image observation, $o_t$, from a downward-looking camera on \ac{ROV} and outputs an action, $a_t$, describing one meter away waypoint consisting of two values, $a_1, a_2 \in [-1, 1]$. The actions represent the direction of the position step and turn in the heading angle, respectively, similar to our previous work \cite{halil}. The reward is defined with respect to vertical divergence from the pipe unless the termination of episodes; $r_t = 10 - 2e_p^2 - 2e_\psi$ where $e_p$ is the closest distance to the pipe in the horizontal plane and $e_\psi$ is the error in heading with respect to pipe direction. An episode is terminated where the pipes are not in the camera's field of view, which is $e_p > 2.5 m$. The \ac{DRL} agent is trained with proximal policy optimization~\cite{schulman2017proximal} algorithm using \texttt{stable-baselines3}~\cite{stable-baselines3} package.

\begin{figure}[t]
    \centering
    \includegraphics[width=8.4cm]{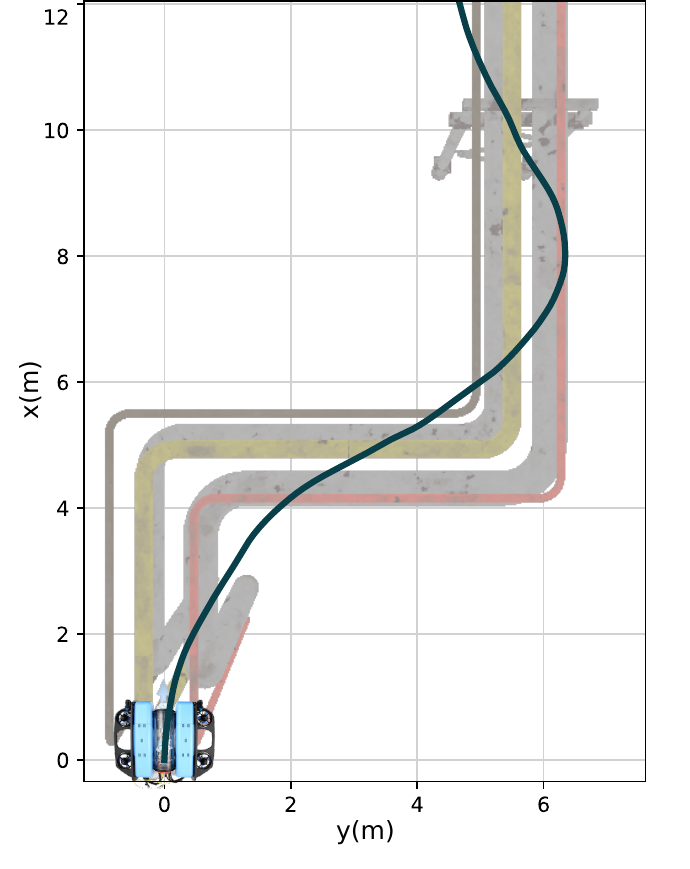}
    \caption{Trajectory followed by the \ac{DRL} agent (dark blue) along with the pipes from the top view. Starting point is the origin of the coordinate frame and is indicated with the \ac{ROV}.
    }
    \label{fig:trajectory_drl}
\end{figure}

The trained policy is deployed on a pipe $\sim20$ meters long with right and left turns. The trajectory of the agent along with the pipes is visualized in Fig. \ref{fig:trajectory_drl}.
While the agent is not accurately tracking the pipes due to the exploratory behavior of the \ac{DRL} policy, it learns to make reasoning from image observations and successfully follows the pipe.
This experiment demonstrates the utilization of high-fidelity image observations and accurate dynamics provided by UNav-Sim in a particular application.

\subsection{Visual localization benchmarking}
The vision-based trajectory generated in Section \ref{sec:example:planning}, being carried out by the robot with the controller showcased in Section \ref{sec:stack:control}, has served as a test setup for the visual localization experiment.
Two trajectories are carried out: a linear trajectory where no area in the map is revisited (see Fig. \ref{fig:trajectory_drl}), and a trajectory with a loop, where the robot navigates back to the starting point.
The benchmarking is automatically performed by the \texttt{robot\_visual\_localization} metapackage using \cite{vSLAM:evo}. 

During runtime, the estimated trajectory $P_i$ and the ground truth trajectory $Q_i$ are recorded into separate files composing a sequence of time-synchronized spatial poses. The pose format is the one proposed in \cite{vSLAM:sturm2012tumrgbd}, composed of the three spatial coordinates with the orientation in quaternions.
The metrics implemented are the \ac{APE} and the \ac{RPE}, which are automatically deployed over the recorded trajectories before shutdown. TartanVO presents a monocular visual odometry algorithm. Therefore, for a fair comparison, the ORB-SLAM3 experiment is executed under a monocular setup. 
The deployment of monocular algorithms implies that the orientation of the algorithm's world frame and the trajectory's scale is arbitrary. Therefore, the estimated trajectories are aligned with the ground truth by obtaining the transform $S \in Sim_3$ that best aligns $P_i$ with $Q_i$.

With the deployment of the automatic stack for visual inspection proposed by UNav-Sim, benchmarking of visual localization algorithms becomes a straightforward task: the robot follows the pipeline autonomously under the planned trajectory, with the visual localization algorithms being automatically executed by the \texttt{robot\_visual\_localization} package, which on shutdown generates the results as depicted in Table \ref{table:comparisonSLAM}. It can be seen from the generated results that the two proposed algorithms depict a similar performance in the linear trajectory, but ORB-SLAM outperforms TartanVO under the presence of a closed loop.
Despite ORB-SLAM's high efficiency in state-of-the-art datasets, the realistic underwater conditions confront one of the main challenges for feature-based approaches: the lack of texture. Moreover, the pipes are the main source of features, which avoids their uniform distribution across the image. Without enough evenly-distributed features, the ORB-SLAM's front end drifts. Nevertheless, the closed trajectory shows the convenience of the back-end's loop closure algorithm: the absolute errors are significantly reduced for translation and rotation.
On the other hand, TartanVO presents a drift similar to ORB-SLAM's in the translations, but slightly higher for rotations. These results show the great potential of learning-based algorithms under imaging conditions that challenge geometry-based methods. Although the lack of generalization ability is the main source of drift in this case, TartanVO has been trained with high amounts of diversified data that explain its good performance in the proposed setup.

In conclusion, the framework proposed in UNav-Sim has enabled the automatic benchmarking of state-of-the-art visual localisation algorithms in a realistic underwater scenario. This has allowed the challenges and opportunities of these algorithms to be easily demonstrated in a geometry-based and learning-based manner.

\begin{table}[ht!]
\centering
\scriptsize
\caption{visual localization results in the pipeline tracking scenario.}
\label{table:comparisonSLAM}
\begin{tabular}{cc c c c c  }
\toprule

Trajectory                   & Algorithm  & APE[m]         & RPE[m]         & APE[rad]       & RPE[rad] \\
\midrule
\multirow{2}{*}[0em]{Linear} & ORB-SLAM3  & 1.75          & \textbf{0.412} & \textbf{1.53} & \textbf{0.036} \\
                             & TartanVO   & \textbf{1.67} & 0.489          & 1.95          & 0.108 \\
\midrule
\multirow{2}{*}[0em]{With loop}   & ORB-SLAM3  & \textbf{0.078}          & 0.372 & \textbf{1.62} & \textbf{0.006} \\
                             & TartanVO   & 0.961 & \textbf{0.322}         & 2.10          & 0.068 \\
\bottomrule
\end{tabular}
\end{table}



\section{Conclusion \& future work} \label{sec:conclusion}
We have developed UNav-Sim, a novel open-source underwater simulator, which builds upon AirSim and \ac{UE5} and incorporates state-of-the-art robotics algorithms. UNav-Sim also supports \ac{ROS} and multiple operating systems, facilitating a streamlined and efficient development process for robotics applications. Future work will include the incorporation of additional underwater sensors, vehicle models, and more custom environments.





\section*{Acknowledgement}
This work is supported by EIVA a/s and Innovation Fund Denmark under grants 2040-00032B and 1044-00007B, the European Union’s Horizon 2020 Research and Innovation Program (OpenDR) under Grant 871449 and the Marie Skłodowska-Curie (REMARO) under Grant 956200. This publication reflects the authors’ views only. The European Commission is not responsible for any use that may be made of the information it contains.



\bibliographystyle{unsrt}
\bibliography{refs}
\end{document}